\pdfoutput=1

\documentclass[11pt]{article}

\usepackage{acl}

\usepackage{times}
\usepackage{latexsym}
\usepackage{pdfpages}
\usepackage{booktabs}
\usepackage[T1]{fontenc}

\usepackage[utf8]{inputenc}
\usepackage{amssymb}

\usepackage{microtype}

\usepackage[]{todonotes}
\makeatletter
\newcommand*\iftodonotes{\if@todonotes@disabled\expandafter\@secondoftwo\else\expandafter\@firstoftwo\fi}  

\everypar{\looseness=-1}

\everypar{\looseness=-1}

\usepackage[]{color-edits}  
\addauthor{Z}{magenta}

\title{The Perspectivist Paradigm Shift: Assumptions and Challenges of Capturing Human Labels} 
\author{Eve Fleisig\textsuperscript{1} \, Su Lin Blodgett\textsuperscript{2} \, Dan Klein\textsuperscript{1} \, Zeerak Talat\textsuperscript{3}\\
\textsuperscript{1}University of California Berkeley \, \textsuperscript{2}Microsoft Research Montr\'{e}al\\ \textsuperscript{3}Mohamed Bin Zayed University of Artificial Intelligence\\
\texttt{\{efleisig,klein\}@berkeley.edu}\\ \texttt{sulin.blodgett@microsoft.com} \, \texttt{z@Zeerak.org}}

\begin{document}
\maketitle
\begin{abstract}
Longstanding data labeling practices in machine learning involve collecting and aggregating labels from multiple annotators. 
But what should we do when annotators disagree? 
Though annotator disagreement has long been seen as a problem to minimize, new \emph{perspectivist} approaches challenge this assumption by treating disagreement as a valuable source of information. 
In this position paper, we examine practices and assumptions surrounding the causes of disagreement---some challenged by perspectivist approaches, and some that remain to be addressed---as well as practical and normative challenges for work operating under these assumptions. 
We conclude with recommendations for the data labeling pipeline and avenues for future research engaging with subjectivity and disagreement.
\end{abstract}

\section{Introduction}
When developing human-labeled data for machine learning (ML) tasks, labels for each example are often obtained by collecting annotations from multiple annotators, which are 
then aggregated to provide a single ground truth label per example.
However, a line of recent work 
has 
illustrated that annotators disagree for many reasons, and that capturing this disagreement can 
improve model performance and calibration \cite{fornaciari_beyond_2021, baan_stop_2022}, 
surface minority voices 
\cite{prabhakaran_releasing_2021}, and 
uncover task ambiguities \cite{balagopalan_judging_2023, parrish_is_2023}. 
Researchers have begun to ask: What should we do when people disagree? 
How can (or should) our datasets and models account for different opinions?\looseness=-1

We argue that this new wave of research---which, following \citet{basile-etal-2021-need}, we refer to as the \emph{perspectivist turn}
---constitutes a paradigm shift in data collection for ML and offers 
an opportunity to systematically examine the changing landscape. 
In this position paper, we examine practices and assumptions across papers 
regarding how 
data is collected from multiple annotators, discuss challenges raised by these approaches, and provide recommendations for rethinking data labeling when annotators disagree. 
We offer our own syntheses of observed practices and assumptions in natural language processing (NLP), 
 as well as observations drawn from meta-analyses of ML 
research more broadly.\looseness=-1 



We first examine what has changed under this paradigm shift: we 
examine 
each paradigm's 
assumptions about the \textit{causes} and \emph{nature} of disagreement, and the \textit{practical challenges} that arise when operating under each 
set of assumptions.
We then explore what has not changed, identifying \textit{normative challenges}---questions and assumptions about labeling not yet taken up in this shifting landscape.
Finally, we offer recommendations for designing data labeling processes 
that better account for annotator disagreement, and avenues for future research. 
By charting these shifting assumptions and practices, we aim to surface the ways in which each paradigm succeeds, or fails, to account for the rich tapestry that disagreement can offer.
\looseness=-1

\section{The Longstanding Paradigm}


We characterize the \emph{longstanding paradigm of data labeling} 
as work that collects labels on a data instance from annotators and aggregates
them with the goal of capturing underlying ground truth labels 
\cite{snow-etal-2008-cheap, nowak2010reliable}.\footnote{Common aggregation strategies include majority vote over labels for binary classification tasks and averaging labels for tasks that use Likert scale ratings. 
We group these under the collective umbrella of ``averaging.''}
By contrast, work in the \emph{perspectivist paradigm} treats variation among annotator labels as a source of meaningful information \cite{basile-etal-2021-need, plank_problem_2022}.
We first examine assumptions about the causes of disagreement and challenges faced under the longstanding paradigm. \looseness=-1

\subsection{What Causes Disagreement?}
\label{sec:causes}


In this section, we examine longstanding practices and assumptions about the causes and nature of disagreement, which the perspectivist paradigm challenges. 
Under the longstanding paradigm, annotator disagreement is often characterized as an issue of label quality, particularly when crowdsourcing labels \cite{nowak2010reliable, artstein2017inter}. 
Disagreement is often attributed to ``subjective,'' confusing, or inherently ambiguous tasks~\cite{aroyo_three_2014}, or to low-quality (inexperienced, uninformed, or biased) annotators \citep{hsueh2009data, nowak2010reliable}. 
Because spam or inconsistency is common, collecting multiple labels per example and measuring inter-annotator agreement can serve as a guarantee of data quality.
\looseness=-1

Perspectivist approaches have re-evaluated several of these practices and their underlying assumptions. 
Here, we discuss three 
such practices: attributing disagreement to bias or ineptitude, requesting labels out of context, and restricting discussion of disagreement to ``subjective'' tasks.\looseness=-1



\paragraph{
Assumption: Disagreement is 
due to biased or inept annotators and thus 
noise to eliminate.} 
In a review of annotator diversity in data labeling, \citet{kapania_hunt_2023} find that ML practitioners ``conflated...
diversity with bias, viewing it...as a source of variability to be corrected or technically resolved'' and attributing it to ``unsatisfying work quality, or worse, questionable work ethics.'' 
Synthesizing previous work, we argue that this assumption stems from 
(1) a conflation between ``bias'' in the statistical sense and societal sense,
and (2) a belief that meaningful differences of opinion only arise due to technical expertise or work quality.
\looseness=-1

\textbf{Disentangling annotator ``bias.''} 
Recent work 
exhibits a conflation between two 
senses of the word \textit{bias}: (i) a statistical sense (as in ``bias-variance tradeoff''), meaning the difference between the expected value of an estimator and its actual value, and (ii) a psychological or societal sense, meaning prejudicial discrimination against a person or group \cite[e.g.,][]{narimanzadeh_crowdsourcing_2023, hube2019understanding, li2020}. 
\citet{kapania_hunt_2023} find that practitioners ``were unable to distinguish minority opinions from `noise' that deviated from instructions.'' 
If the mean label $m$ of a group of annotators is considered the ground truth for a data example, then an annotator whose label is far from $m$ is statistically biased. 
Yet, we argue, it does not follow that the annotator must be societally biased: for example, if the annotator is a member of an affected community who knows more than other annotators about the context of the example being labeled, it may instead be the mean label that is societally biased.
Since disagreement manifests as statistical bias, which is equated with societal bias, all disagreement is undesirable under this assumption.

\textbf{Disentangling ``expertise.''} 
Though machine learning 
acknowledges the value of ``expert annotators''---generally people with prior training in an area, or quantifiable knowledge such as fluency in a language---\citet{kapania_hunt_2023} find that annotators are rarely recruited ``based on their lived experiences, knowledge, or expertise as facets of diversity.'' 
When lived experience is not seen as a legitimate source of expertise,  disagreement on that basis 
is more easily ascribed to ``bias'' than well-informed but different views. 
Multiple studies have indeed found that annotator opinions vary based on factors related to lived experience, including  demographics, political views, and community membership 
\cite{Patton2019AnnotatingSM, larimore-etal-2021-reconsidering, sap-etal-2019-risk}. 
These findings indicate 
that lived experiences shape people's judgments, and therefore that ``non-experts'' with different backgrounds can disagree 
without being ``low-quality'' annotators. 
In turn, this suggests that such disagreement ought to be treated as meaningful in its own right.\looseness=-1


\paragraph{Practice: Annotators rarely receive task context.}
Data labeling tasks often give annotators minimal context when labeling data~\cite{Fortuna_Directions_2022}, thus implicitly treating such context as irrelevant to annotators' decision making. 
Nevertheless, context can greatly change annotator behavior; for example, in hate speech detection, giving annotators context about text authors' probable race or language variety changes annotator judgments 
\cite{sap-etal-2019-risk}, while in machine translation, detailed instructions increase annotator agreement \cite{popovic-2021-agree}. 
Information given to annotators about how labels will be used also affects their judgments, even with no change in the data being labeled:
\citet{balagopalan_judging_2023} ask annotators to do the same task framed as a factual classification or as a judgment of whether a norm was violated (e.g., 
whether an outfit matches a description or breaks a dress code based on the same description), and find that annotators are ``less likely to say that a rule has been violated than to say that the relevant factual features...are present.'' 
This suggests that annotators account for potential consequences that are salient to them---e.g., penalizing people for breaking a dress code. Thus, annotators' assumptions about task context---which under the longstanding paradigm have typically remained implicit---may represent an overlooked source of meaningful disagreement.\looseness=-1 

Indeed, recent work has indicated that annotators are aware of the impact of the assumptions they make on decontextualized tasks, and sometimes request more granular instructions and context. 
Surveying Mechanical Turk workers on the types of information that help on confusing tasks, \citet{huang2023incorporating} find that over 50\% of annotators want more context on annotations, and over two-thirds believe that knowing the purpose of the labeling task would help them.

\paragraph{
Assumption: Disagreement is 
limited to ``subjective'' tasks.}
It is tempting to assume that disagreement is limited to tasks based on personal opinions, such as those that involve the quality of art or text, or those that touch on sociocultural norms, such as offensive speech detection. 
Yet disagreement arises even in seemingly clear-cut tasks, such as natural language inference (NLI) \cite{pavlick-kwiatkowski-2019-inherent,jiang_investigating_2022} and semantic textual similarity \cite{wang_collective_2023}. 
\citet{geva_are_2019} find that responses on NLI and question answering tasks vary enough by person that annotator-specific models improve downstream task performance, while 
\citet{parrish_is_2023} find that in image classification, issues such as differing names for the same objects in different regions and differing interpretations of a task (e.g., whether a picture of a bird counts as a ``bird'') result in disagreement. 
\citet{basile-etal-2021-need} note other causes of annotator disagreement, such as task complexity, annotator proficiency at the task, and cognitive biases. 
These varied factors suggest there is no clear set of tasks that admit no subjectivity or disagreement.\looseness=-1

\subsection{Practical Challenges under the Longstanding Paradigm}
\label{sec:pop-opinions}


Having examined this paradigm's assumptions about the causes of disagreement, in this section we accept its goal---to capture a single underlying ground truth annotation per example, ideally the broader population's opinion
---at face value and examine technical challenges towards achieving it in practice. 
We argue that even under the longstanding assumption that capturing such a label is possible and that annotator disagreement does not reflect meaningfully differing opinions, a number of technical challenges across different stages of data labeling continue to make capturing labels difficult. Specifically, we suggest that \textbf{collected labels are not a good proxy for the stakeholder population's views}, and that 
\textbf{diverse recruitment is not enough}, because even uniform sampling of the annotator pool with aggregated labels inaccurately models the broader population for several reasons:\looseness=-1

\textbf{Unrepresentative annotator pools.} 
The demographics of 
crowdworking platforms such as Mechanical Turk 
are not representative of most populations of interest (including system users, affected stakeholders, or even the population of the regions from which crowdworkers are recruited).
For example, U.S. Mechanical Turk workers are disproportionately white and young compared to the general U.S. population \cite{PewMturk}.\looseness=-1 

\textbf{Sample error.} 
When small numbers of annotators are recruited relative to the population size, the average of their ratings is likely to be farther from the average of the full population \cite{narimanzadeh_crowdsourcing_2023, geva_are_2019}. 
 This effect is exacerbated 
when few annotators annotate each data item, making it less likely that an annotator with relevant background is assigned to annotate a particular item. 
Moreover,
under current crowdsourcing practices, there is 
often no limit on how many annotations one person may do, resulting in datasets that may reflect only the opinions of the most prolific annotators \cite{geva_are_2019}.

\textbf{Aggregation treating 
minority opinions as noise results in miscalibrated models.} 
\citet{narimanzadeh_crowdsourcing_2023} note that majority voting always discards data from ``minority raters holding less popular opinions'' and moves the estimated mean further from the true population mean by non-randomly discarding ratings.
Aggregated judgments have disproportionately high agreement with white annotators \cite{prabhakaran_releasing_2021}, reflecting the fact that aggregated labels typically reflect the opinions of groups with higher representation and minimize the representation of minority opinions.
As a result, downstream models are often miscalibrated with respect to diversity of opinions between annotators \cite{baan_stop_2022}.

\subsection{Normative Challenges}
\label{sec:underlying}
While the perspectivist literature has identified and challenged a number of longstanding assumptions about disagreement, several longstanding assumptions remain only partially addressed even in perspectivist work. 



\paragraph{Sometimes, there is no ground truth.}
The existing paradigm of data labeling implicitly imagines annotation as a process of uncovering the single ``ground truth'' label for the data, using annotators as noisy approximators. 
However, findings across a range of tasks suggest that there is often no such ground truth. This may occur because the task is underspecified (e.g., the intent of the data labeling process is not clear enough to the annotators to eliminate all ambiguity); the fact that disagreement occurs even in tasks not usually seen as ``subjective'' highlights the difficulty of removing all potential ambiguity. Alternatively, it may occur because reasonable people who fully understand the intent of the annotation could have different opinions, leaving the ``ground truth'' undefined. 


\paragraph{Averaging labels loses information about a population's values.} 
Averaging opinions, e.g., 
via majority vote, has a millennia-old history as a way of democratically aggregating views on an issue \cite{boegehold-voting}.
However, naively averaging data labels encounters serious issues in practice.
People are not equally well-informed or culturally grounded for all tasks, nor do they face equal consequences from model decisions. 
Expertise---including less quantifiable factors such as lived experience and sociocultural background---is key for many tasks, particularly when a task affects a particular community. 
Yet averaged labels ignore such 
considerations, resulting in lower-quality datasets that may disregard 
those who are most affected. 

\section{The Perspectivist Turn}
\label{sec:perspectivism}
Perspectivist efforts argue that longstanding approaches are insufficient when 
(1) annotators frequently disagree in ways that are 
important to capture
, and (2) even with diverse annotator recruitment, aggregate labels often fail to adequately represent the true population's opinions. 
Approaches in the perspectivist turn include training with annotators' individual labels or pertinent details about the annotators 
and explicitly modeling individual annotators' behavior \cite[e.g.,][]{davani_dealing_2022, gordon_jury_2022, plepi_unifying_2022, sachdeva-etal-2022-measuring}; training with probability distributions over labels \cite{fornaciari_beyond_2021, uma_case_2020}; calibrating to variance between annotators \cite{baan_stop_2022}; collecting labels from many annotators \cite{nie-etal-2020-learn, aroyo2023dices}; and investigating causes of disagreement \cite[e.g.,][]{goyal_is_2022, larimore-etal-2021-reconsidering, pei_when_2023}.\footnote{See \citet{plank_problem_2022} and \citet{basile_2023} for 
discussions of the range of perspectivist work.} 
Here, we explore how perspectivist approaches conceptualize the causes and nature of disagreement, as well as emerging practical and normative challenges.

\subsection{Rethinking Causes of Disagreement}
\label{sec:causes-perspectivist}

Perspectivist approaches have challenged many, but not all, of the longstanding assumptions described in Section \ref{sec:causes}. In this section, we chart how these approaches reconceptualize disagreement.\looseness=-1

\paragraph{Perspectivist approaches recognize that annotator demographics and lived experiences can result in disagreement.} Recent studies have examined demographic factors that lead to disagreement, such as race, gender, and age, as well as cultural factors such as education, political affiliation, and native language proficiency (e.g., \citealp{goyal_is_2022, jakobsen_being_2023, al_kuwatly_identifying_2020, wan_everyones_2023, pei_when_2023}), with a view toward ensuring that the opinions of people from different backgrounds are represented. 

\paragraph{Nevertheless, differences between demographic groups only partly explain disagreement.} While this work has been important in better understanding where and how disagreement arises, these methods often assume that disagreement can be well-characterized 
by demographic factors alone. 
However, recent work suggests that non-demographic factors 
are more probable sources of disagreement than some demographic factors across multiple tasks. Many demographic factors do not appear to be good predictors of disagreement across all tasks;
\citet{orlikowski-etal-2023-ecological} find that modeling gender, age, education, and sexual orientation in isolation do not predict disagreement effectively on a hate speech task, \citet{biester-etal-2022-analyzing} find no significant differences based on gender across multiple tasks, and \citet{fleisig_when_2023} find that while race is an important factor in predicting disagreement on hate speech detection, factors such as gender and education are not.

Conversely, factors beyond demographics often cause differences in opinion. 
These may be task-specific; for example, social media usage and opinions on whether online toxic content is a problem greatly help to predict labels on hate speech detection \citep{fleisig_when_2023}. 
Other key factors lie outside the scope of what perspectivist work has considered. 
For example, \citet{miceli_posada_2022} describe ``errors'' by Venezuelan image labelers due to differences between English and Spanish, since translations of some words refer to slightly different set of objects. 
Such issues suggest that a wide range of experiences and perspectives not well-captured by demographics may help to explain systematic disagreement between annotators, but only some of these have been explored.\looseness=-1 


Regardless of the predictive power of demographics, understanding the opinions of stakeholders from a range of demographic backgrounds is a key contribution of perspectivist work: 
both because it is important that people from a range of different backgrounds be heard even if they often agree, and because views on more specific topics can vary along demographic axes even if they are not relevant for every item in a dataset. 
However, widening the scope of potential causes of disagreement would deepen our understanding of why disagreement occurs, improve modeling of annotator behavior, and help to target annotator recruitment to axes that cause disagreement for specific tasks.

\subsection{Emerging Practical Challenges}
\label{sec:tasks-perspectivist}
Perspectivist approaches have re-evaluated many longstanding assumptions in data annotation regarding the origins and value of disagreement. 
However, its new ambitions to engage with the full spectrum of human perspectives bring new challenges regarding data quality, data ethics, institutional pressures, and personalization.

\paragraph{Assessing data quality while capturing disagreement is difficult but critical.} A major motivating factor for aggregating multiple annotators' labels is the concern over spam and inattentive or inept annotators, resulting in much research focused on maximizing agreement as a metric of data quality (see Section \ref{sec:causes}). 
The tension between preserving all annotator opinions and removing ``noise'' means that perspectivist approaches will face limited use unless alternative methods are developed to maintain data quality without discarding disagreement. 
Promising examples of these methods include \citet{Marneffe2019TheCI}, which uses clear-cut control samples for which the authors are willing to assume that no disagreement could reasonably occur. \citet[Appendix B]{deng-etal-2023-annotate} collect a variety of quality checks from previous work that, besides inter-annotator agreement, include completion time \cite{diaz2018}, correlation between similar labels \cite{demszky-etal-2020-goemotions}, and briefing or training annotators \cite{akhtar2021opinions}.

\paragraph{Evaluation still relies primarily on majority-vote labels.} 
\citet{plank_problem_2022} notes that a majority of perspectivist papers still evaluate against averaged ``gold'' labels, which undercuts the potential utility of perspectivist methods. 
We argue that the continued evaluation via averaging is a symptom of deeper problem:
even if we model diverse annotator opinions, models typically produce a single output or classification, and we lack metrics for the quality of that single output besides its similarity to the gold aggregated label. 
That is, despite the more detailed and diverse data gathered from perspectivist work, the community lacks methods to evaluate models using that data (though see Section \ref{sec:recommendations} for approaches beginning to explore such methods).

\paragraph{Collecting more detailed data requires considering impacts on data subjects.}
Collecting the opinions of minoritized populations could constitute an undue burden on minoritized groups, especially if the data collection does not result in a commensurate benefit in terms of quality of service for that group. 
There is also a tradeoff between the richness of collected data and preserving privacy of group members. 
Potential ways forward include learning from less data so fewer data points are needed, using privacy-preserving machine learning methods \cite{Xu2021PrivacyPreservingML}, and engaging with community-led methods for preserving data ownership, such as indigenous data sovereignty \cite{kukutai_taylor}.

\paragraph{Participatory approaches conflict with institutional pressures.} 
Institutional pressures hinder efforts to collect more representative and complex data, particularly when it comes to meaningfully involving participants. 
Researchers face pressures to collect data quickly, not better. 
By contrast, participatory approaches aim to build mutual, reciprocal relationships; grapple explicitly with power dynamics between researchers and participants, as well as between participants; engage with specific contexts of use; and rethink what is on the table for participants---for example, extending beyond data collection to problem formulation or evaluation \cite{delgado2023}.
Thus, calls to increase participation 
may underestimate the extent to which institutional factors discourage such approaches. 
As a result, lowering boundaries to participation through platforms or methods of data labeling that improve communication and empower participants is key, as well as pressuring institutions to incentivize slower, more thoughtful, and more context-specific (rather than maximally portable \cite{Selbst2019}) data collection.\looseness=-1 

\subsection{Emerging Normative Challenges} 

Perspectivist work exposes longstanding assumptions regarding ground truth and the merits of aggregation, but some assumptions still remain implicit in perspectivist approaches. 
We delineate 
normative challenges that perspectivist work still faces regarding majority-vote labels, the bounds of acceptable disagreement, and researcher positionality.\looseness=-1 

\paragraph{
Perspectivist approaches do not always explicitly take a normative stance.}
Machine learning researchers often do not take explicit stances on what systems ought or ought not to do, under the assumption that research is or should be neutral and does not reflect social values or researcher perspectives \cite{Birhane2022,santy-etal-2023-nlpositionality}. But as emerging perspectivist efforts aim to engage with the full spectrum of human perspectives, researchers and practitioners will need to grapple explicitly with challenging normative questions---does the problem formulation admit a correct answer, and (if there is one) whose perspectives form the basis for that answer? Are some perspectives prioritized, or they are all weighed equally? 

Engaging explicitly with these questions is especially critical because not doing so may leave important assumptions implicit and therefore unavailable for discussion \cite{blodgett-etal-2020-language}, or even cause its own harm \cite{Talat_Disembodied_2021}. For example, in the absence of explicit definitions of hate speech, research may instead rely on aggregation of crowdsourced perspectives to decide what constitutes hate speech. But such an aggregation may in fact unjustly neglect the views of minoritized groups \cite{thylstrupDetectingDirtToxicity2020}.

We therefore see discussion of these normative questions as essential as the perspectivist literature continues to develop. If, as we suggest in Section \ref{sec:tasks-perspectivist}, the community ought to be developing the technical machinery to model and evaluate beyond majority vote labels, then as a prerequisite, the community must explore what it wants that machinery to model and evaluate.

\paragraph{Bounds of ``acceptable'' disagreement typically remain implicit.}
\label{sec:norms-perspectivist}
\citet{rottger-etal-2022-two} distinguish between a descriptive annotation paradigm, in which annotators are encouraged to provide subjective opinions without researcher influence, and a prescriptive one, in which annotators are encouraged to be ``objective'' and adhere to strict guidelines. 
This dichotomy can aid researchers in deciding whether disagreement on a data labeling task should serve as a signal that the task is underspecified or as valuable information to preserve. 

Many data labeling tasks combine descriptive and prescriptive practices. 
Task-specific bounds of acceptability often define when variation should be preserved: a painting of a bird might be reasonably labeled ``painting'' or ``bird,'' but not ``cat.'' 
Setting these bounds is particularly fraught for tasks involving social norms, such as hate speech detection. 
Understanding where to set guidelines, and where to permit variation, is task-specific and difficult. For example, there is widespread disagreement over how to operationalize ``toxicity'' and ``alignment,'' concepts whose bounds often go unstated despite being central to major ``subjective'' tasks \cite{thylstrupDetectingDirtToxicity2020, kirk2023signifier}. 
However, without explicitly setting such bounds, we encounter the problems faced under the majority-vote paradigm: opinions defined nebulously by aggregation result in normative boundaries that are hard to pinpoint, let alone contest. 
These boundaries may thus be difficult to change even when they are demonstrably unfair.\looseness=-1

\paragraph{Personalization may not resolve issues of disagreement.} 
Increasingly powerful language models present the possibility of  
personalizing models to individual users rather than using a single model to satisfy many different preferences  \cite{plepi_unifying_2022, flek_returning_2020}. 
We argue that personalization alters issues related to disagreement but does not 
necessarily solve them. 
While some types of personalization are beneficial (e.g., targeting a scientific explanation to students at different levels), others could perpetuate harms (e.g., supporting misinformation that a user believes). 
Personalization does not bypass normative issues, but rather changes the structure of the problem: the difficult decision becomes whether and when personalization is appropriate. 
Here, the community might draw on work in recommender systems, in which personalization is a primary goal and persistent concerns arise about its appropriate scope and potential harms \cite{ekstrand2012fairness, Wang2022ASO, Li2023}.


\section{Recommendations for Practical Challenges} \label{sec:recommendations}
Annotator disagreement carries implications for all stages of the data labeling process. 
We provide recommendations for each of these stages:

\paragraph{Before data labeling begins.} 
If prescriptive decisions are made about acceptable bounds of disagreement when designing a data labeling process, these decisions should be made explicitly. 
In addition, consider 
potential axes of disagreement for the specific task at hand, such as linguistic or sociocultural differences, ambiguous labels, or differences of opinion. 
Considering these normative questions---who or what is the data collection for---and potential sources of disagreement before beginning data labeling can help to design the process so that important differences in opinion are captured, and sources of confusion are minimized.\looseness=-1

\paragraph{Recruitment.}
Depending on the extent to which the collected data should reflect the opinions of all stakeholders or focus on experts, different best practices for annotator recruitment apply. If the objective is to reflect the views of a particular population, such as potential users, it is crucial to recruit a representative sample of that group. This may sometimes require additional recruitment efforts to account for different demographics' uneven participation in crowdsourcing. In addition, rather than filtering out ``noisy'' annotators based on whether they disagree with others, alternative filtering strategies such as checking intra-annotator agreement \cite{abercrombie_consistency_2023} or doing multiple rounds of qualification tasks before the main task \cite{Zha:22} can help to reduce spam without discarding minority opinions.

An intermediate approach might consider stratifying the recruited sample of annotators based on important axes of disagreement (e.g., different countries where a model will be used) to upsample  groups that might otherwise be underrepresented.
In addition, for tasks involving different types of expertise (e.g., system summarizing medical or legal documents, or a language model giving advice to specific communities), consider allocating annotators to items based on their expertise.\looseness=-1

Other considerations apply regardless of the recruited population. Recruiting a large annotator pool helps mitigate sample error, and capping annotations per annotator can prevent a dataset from primarily reflecting the views of a few annotators. When modeling disagreement, consider collecting annotator data specifically about factors likely to cause disagreement for the task at hand.\looseness=-1

\paragraph{Data labeling design.}
Given annotators' frequent concerns over a lack of task context, and the effects of task context on annotator judgments, it is key to give annotators more context when labeling data. This includes what the data will be used for (e.g., for what task, for which users) and potential effects of system decisions (e.g., whether the system will be used in a punitive way). Furthermore, use disagreement as a signal to prompt reflection and iteration on the data labeling process. For example, disagreement can signal confusing instructions or an insufficiently rich space of potential labels. In cases where ambiguities could cause disagreement (e.g., whether pictures of birds count as birds), or where annotators might provide labels not foreseen by task designers \cite[e.g.,][]{sheppard2023subtle}, provide ways for annotators to indicate uncertainty, such as an ``unsure/unclear'' option, and ways to give open-ended feedback so that the task can be clarified or expanded.
\looseness=-1

\paragraph{Dataset documentation.}
Details on the data labeling process can help future stakeholders to understand factors that might have affected annotator judgments. Previous data documentation work has recommended including information such as annotator demographics and labeling task instructions \citep{mcmillan-major} or the original task for which data was collected \citep{gebru2021datasheets}. Expanding on this work, we recommend also documenting (i) annotator selection procedures, including the number of annotators and restrictions on participation, (ii) the distribution of items labeled per annotator, and (iii) any annotator filtering used. Future dataset users can also benefit from describing normative bounds imposed on the data labeling process and rationales for discarding any data. Providing non-aggregated individual labels when possible also helps to avoid information loss from aggregation.\looseness=-1

\paragraph{Model design and evaluation.}
Different model objectives aside from accuracy on predicting aggregate labels, such as measuring KL divergence between predictions and the distribution of annotator labels, or calibrating to the distribution of annotator opinions, allow disagreement to be accounted for during training. During evaluation, potential alternatives to using averaged ``gold'' labels include measuring distributional similarity, e.g., with KL divergence, cosine similarity between lists of outputs, or a correlation coefficient \cite{nie-etal-2020-learn, dumitrache-etal-2019-crowdsourced, zhou-etal-2022-distributed}, evaluating accuracy at modeling individual annotators \cite{davani_dealing_2022, resnick_2021}, and measuring model calibration to population uncertainty \cite{baan_stop_2022}. 
Evaluator disagreement is also a useful signal: if evaluators disagree over the quality of a model output, this information can help to pinpoint model weaknesses or reveal instances of disparate quality of service for different subgroups.\looseness=-1

\section{Recommendations for Normative Challenges}
\label{sec:forward}
In this section, we discuss potential avenues for research aiming to engage annotator disagreement.



\paragraph{Replace implicit normative decisions with explicit ones.} 
Majority-vote aggregation captures the average view of the aggregated population with every annotator weighted equally. 
By contrast, data labeling tasks where some people are clearly better-informed (e.g., doctors in medical domains, or speakers of a language for translation) have an implicit ``expert-driven'' framing, in which only some views are solicited. This includes considering lived experience as a form of expertise, which can prove critical to successful annotation. 
We can imagine a spectrum of practices ranging from ``democratic'' to ``expert-driven,'' with different points along this spectrum suited to different situations. 
For a task requiring 
medical knowledge, it would be unreasonable to use labelers with no medical training; when setting community norms, all community members' views are important. 
Each data labeling task requires choosing a point on this spectrum. 
Making this decision 
explicitly, rather than defaulting to majority vote, can help to create decision rules that are easier to define and contest.

\paragraph{Draw on parallel problems from other disciplines.}
The broader questions of how to capture a population's views, and make decisions based on them, has a long history across a range of traditions involving stakeholder participation, from social choice theory and mechanism design \cite{Arrow1977-dh,Feldman2006-cy} to value-sensitive and participatory design \cite{friedman1996value, muller1993participatory}:

\textbf{Science and technology studies.}
\citeposs{Bowker2000} analysis of the political and social dimensions of classification highlights the importance of retrievability, the process of retaining the voices of people conducting classification for systems to maintain ``maximum political flexibility.'' As perspectivist methods examine ways to retain the opinions of individual labelers, this line of work can help to understand how individual voices can be lost, merged, or preserved in systems that draw on them; and understand how we can, as \citet{Bowker2000} note, ``reflect new institutional arrangements or personal trajectories.'' \citeauthor{Bowker2000}, as well as \citet{Winner1980} and \citet{agre2014toward}, illustrate how technological artifacts embed and reproduce social and political values; drawing on \citeauthor{douglas_purity_1978}, \citet{Lepawsky2019} 
and \citet{Scheuerman2021} investigate institutional factors and perspectives of researchers or industry leaders who describe subjectivity as a problem to minimize. Together, this literature contextualizes  subjectivity and disagreement, and emphasizes the need for critical reflection on practices and assumptions in technological development.


Elsewhere, critiques of machine learning, including approaches to fairness and ethics, can offer opportunities for perspectivist efforts to reflect on assumptions surrounding representation and inclusion. For example, \citet{Hoffmann2020} complicates the notion of inclusion in dataset design by pointing out that such inclusion can forestall calls for more radical change, while \citet{Stevens2021} similarly observe that more representative datasets for e.g., facial recognition do not address the more fundamental problem of surveillance.\looseness=-1


\textbf{Philosophy of mind.} Literature related to why, despite our understanding of the physical processes involved, we still lack a full understanding of where subjective feelings come from and why they differ, such as discussion of qualia \cite[among others]{Lewis1930-LEWMAT-10, Jackson1982, Chalmers1997-mj}, can provide a starting point for discussion of differences of opinion that are not easily situated in terms of the annotator’s background.

\textbf{Voting and social choice.} Many issues regarding optimal data labeling resemble issues regarding ideal voting mechanisms, with different constraints. 
In electoral settings, the full population's opinions may be solicited, and single decisions based on their choices have widespread effects (e.g., electing an official who makes decisions in many policy areas). 
During annotation, by contrast, it is 
often infeasible to solicit 
all stakeholder opinions, and different aspects of model outputs can be decided independently (e.g., decisions on coherence or offensiveness of model outputs). 
However, overarching themes of how to aggregate preferences while maximizing 
stakeholder satisfaction and welfare \cite{Arrow1977-dh, sen2018collective} could provide useful lessons for perspectivist work.


\textbf{Pragmatics.} The community could take inspiration from the notion of a ``common ground'' in pragmatics, wherein conversational participants communicate based on a shared understanding of the world. This shared understanding is based on factors that include demographic attributes, but also factors such as the specific speech situation, the participants' professions, online communities, languages spoken, and imagined audiences \cite{clark1982, goffman_1976}. Annotation of text functions like a communicative situation in which the annotator interprets language while making assumptions about the speaker, purpose, and audience of the text based on their own background, and a wide variety of factors in their background may be relevant based on the text. Focusing on content moderation, \citet{thylstrupDetectingDirtToxicity2020} draw on \citet{hall1997spectacle} to describe how these assumptions become embedded in datasets:  data annotators serve as intermediaries who read on behalf of the intended recipient and often interpret text differently from the specific intended reading that the sender meant to encode, with the result that systems based on those labels encode the intermediary position instead of that of the sender or intended recipient. Understanding the range of factors that influence annotator interpretation could disentangle more latent factors behind disagreement in data collection.

\textbf{Participatory design.} Elsewhere, participatory traditions that interrogate power dynamics in order for ``non-expert stakeholders to provide direct input on technology design'' \cite{delgado2023} can offer practitioners valuable insights for navigating disagreement and 
reflecting on assumptions about disagreement embedded in their practices
\cite{friedman1996value, muller1993participatory}.

\paragraph{Take advantage of nuanced output spaces to meet diverse stakeholder needs.}
The existence of disagreement does not rule out the possibility of building systems that produce single outputs affecting a whole population. \citet{edenberg_disambiguating_2023} note that ``any society that protects freedom of thought and expression'' experiences ``continued disagreement about key normative questions,'' but we still ``find fair terms of social cooperation without requiring everyone to agree.'' 
Even when providing a single output is unavoidable, treating preferences non-unidimensionally can help to arrive at single outputs that  better serve more people. Systems that provide for a broad range of potential outputs, including generative models, can help to consider different, non-contradictory values that seem to result in disagreeing preferences. For example, if one annotator prefers a language model output that is non-discriminatory and another prefers one that is concise, they might disagree on their preferences between two outputs, but a non-discriminatory and concise output could satisfy both annotators. 
Revealing that preferences are not unidimensional and exploring the resulting space of potential  outputs opens ways to generate greater consensus.\looseness=-1

\section{Conclusion}
Assuming that tasks have a ground truth, using majority-vote aggregation, and avoiding a normative stance have long been common practices in data labeling. However, a growing perspectivist literature is recognizing that datasets and models must be designed to account for the full spectrum of human perspectives. 
 We argue that perspectivist approaches can accomplish their goals more fully by considering causes of disagreement beyond demographics, addressing tensions with data quality and research pressures, and reasoning explicitly about normative considerations. 
\looseness=-1

\section*{Limitations and Ethical Considerations}
Our position paper aims to provide an analysis of key questions regarding longstanding and emeerging paradigms of data collection, but it is not a comprehensive meta-analysis or literature review; thus, we acknowledge that some relevant work may have been overlooked because we have not comprehensively searched for all papers related to these issues. Overlooking some work carries the risk of narrowing the set of potential perspectives that are considered in future research based on the avenues we discuss.

\section*{Acknowledgments}
Thank you to members of the Berkeley NLP and Algorithms, Data, and Society groups for their feedback, particularly Deborah Raji and Nicholas Tomlin. Thank you as well to the anonymous reviewers for their helpful suggestions.


\bibliography{anthology,custom}
\bibliographystyle{acl_natbib}

\appendix
\end{document}